\documentclass{article}
\usepackage{ijcai21}

\usepackage{times}
\usepackage{soul}
\usepackage{url}
\usepackage[hidelinks]{hyperref}
\usepackage[utf8]{inputenc}
\usepackage[small]{caption}
\usepackage{graphicx}
\usepackage{amsmath}
\usepackage{amsthm}
\usepackage{booktabs}
\usepackage{algorithm}
\usepackage{algorithmic}
\usepackage{array}
\usepackage{colortbl}
\usepackage{multirow}
\usepackage{tablefootnote}
\usepackage{balance}
\newcommand{\tabincell}[2][c]{\begin{tabular}[#1]{@{}c@{}}#2\end{tabular}}

\title{Knowledge-aware Zero-Shot Learning: Survey and Perspective}

\author{
Jiaoyan Chen$^{1*}$
\and
Yuxia Geng$^2$\and
Zhuo Chen$^{2}$\and
Ian Horrocks$^1$ \and
Jeff Z. Pan$^3$ \And
Huajun Chen$^2$\thanks{Corresponding authors} 
\affiliations
$^1$Department of Computer Science, University of Oxford\\
$^2$College of Computer Science \& AZFT Knowledge Engine Lab, Zhejiang University\\
$^3$School of Informatics, The University of Edinburgh
}

\begin{document}

\maketitle

\begin{abstract}
Zero-shot learning (ZSL) which aims at predicting classes that have never appeared during the training using external knowledge (a.k.a. side information) has been widely investigated.
In this paper we present a literature review towards ZSL in the perspective of external knowledge, where we categorize the external knowledge, review their  methods and compare different external knowledge.
With the literature review, 
we further discuss and outlook the role of symbolic knowledge in addressing ZSL and other machine learning sample shortage issues.
\end{abstract}

\section{Introduction}
Normal supervised machine learning (ML) classification trains a model with labeled samples and predicts the classes of subsequent samples using classes that were encountered during the training stage.
\textit{Zero-shot learning} (ZSL), however, aims to also predict novel classes that did not occur in the training samples.
Such novel classes are known as \textit{unseen classes}, while the classes occurring in training samples are known as \textit{seen classes}.
ZSL has been widely investigated as a means of addressing common ML issues such as emerging classes, sample shortage, etc.

ZSL was originally proposed for image classification in Computer Vision (CV) \cite{palatucci2009zero,lampert2009learning}.
One typical case study is recognizing animals that have no training images by exploiting their semantic relationships to animals that have training images through external knowledge (a.k.a. \textit{side information}) 
such as text descriptions, visual annotations and the taxonomy. 
It has since been applied to Natural Language Processing (NLP) tasks such as text classification and relation extraction (a.k.a. relation classification), as well as to ML tasks in other domains such as Knowledge Graph (KG)~\cite{Pan2016} construction and completion.
Until now hundreds of papers have been published concerning ZSL theories, methods and applications.

Although several review papers have been published, they mostly focus on categorizing the ZSL settings and the algorithm design patterns (e.g., into classifier-based and instance-based) \cite{fu2018recent,wang2019survey,xian2018zero}, and few of them systematically analyze the external knowledge which play a key role in designing ZSL methods and in improving their performance since no samples are given for the unseen classes.
In contrast, this paper reviews ZSL studies mainly from the perspective of external knowledge.
We categorize the external knowledge into four kinds --- text, attribute, KG and ontology \& rules, according to their data structures, sources, expressivity, etc.
For each kind we review its methods, case studies and benchmarks.
We also compare different external knowledge, and discuss the role of symbolic knowledge representation in addressing ZSL and other sample shortage settings. 
Although \cite{fu2018recent} and \cite{wang2019survey} also introduce the semantic space (i.e., encoding of the external knowledge), no paper analysis is conducted for each external knowledge, and more importantly the external knowledge involved (mainly text and attributes) are quite incomplete --- KG and ontology which started to be widely investigated in recent three years are not covered.
\cite{xian2018zero} contributes a comprehensive evaluation to multiple ZSL methods, but these methods are limited to image classification, while we consider different tasks in multiple domains including CV, NLP, KG construction and completion, etc.


\section{Overview}\label{sec:overview}

\subsection{Problem Definition and Annotations}

In ML classification, a classifier is trained to approximate a target function $f:x\rightarrow y$, where $x$ represents the input data and $y$ represents the output class.
\textit{Standard ZSL} aims to classify data with the candidate classes that have never appeared in the training data.
We denote \textit{(i)} the training samples as $\mathcal{D}_{tr} = \{(x, y) | x \in \mathcal{X}_s, y \in \mathcal{Y}_s\}$ where $\mathcal{X}_s$ and $\mathcal{Y}_s$ represent the training sample inputs and the seen classes, respectively;
\textit{(ii)} the testing (prediction) samples as $\mathcal{D}_{te} = \{(x, y) | x \in \mathcal{X}_u, y \in \mathcal{Y}_u\}$ where $\mathcal{X}_u$ and $\mathcal{Y}_u$ represent the testing sample inputs and unseen classes, respectively, with $\mathcal{Y}_u \cap \mathcal{Y}_s = \emptyset$;
and \textit{(iii)} the external knowledge (side information) as a class semantic encoding function $h: y \rightarrow z$ where $z$ represents the semantic vector of the class $y$, $y \in \mathcal{Y}_u \cup \mathcal{Y}_s$.
Note the external knowledge are originally represented as symbolic data such as class names, class textual descriptions and inter-class relationships.
To be involved in ZSL, they are transformed into sub-symbolic representations (i.e., vectors).
The ZSL problem is to predict the classes of $\mathcal{X}_u$ as correctly as possible.
Specially, when the candidate classes in $\mathcal{D}_{te}$ are set to $\mathcal{Y}_u \cup \mathcal{Y}_s$, the problem is known as \textit{generalized ZSL}.

Note in addressing some tasks such as KG link prediction, the original modeling function $f$ is often transformed into a scoring function by moving $y$ to the input, denoted as $f':(x,y)\rightarrow s$ where $s$ is a score indicating the truth degree of $y$.
With $f'$ the class of a testing sample $x$ in $\mathcal{X}_u$ is predicted as the class in $\mathcal{Y}_u$ (or $\mathcal{Y}_u \cap \mathcal{Y}_s$) that maximizes $s$. 
%
Considering a link prediction task with two given entities as the input $x$ and one relation as the class $y$ to predict, the score $s$ quantifies whether the relation matches the two entities' relationship.

\subsection{Technical Solutions}\label{sec:ts}
With the external knowledge and the class semantic encoding function $h$ (cf. details in Section \ref{sec:ek}), a few machine learning methods can be applied to address the above ZSL problem.
We divide the majority of them into three kinds.

\subsubsection{Mapping Function Based}
Given the training samples $\mathcal{D}_{tr}$, the mapping function based methods learn a mapping function from the input space to the class semantic encoding space, denoted as $m: x \rightarrow h(y)$.
In prediction, they adopt the nearest neighbour searching in the class semantic encoding space.
Namely, for a testing sample $x'$ in $\mathcal{D}_{te}$, its predicted class is calculated as $\arg \min_{y \in \mathcal{Y}_u} d(m(x'), h(y))$, where $d$ is a metric that calculates the distance between two vectors, such as the Euclidean distance.
Some methods such as \cite{frome2013devise} learns a linear mapping function,
while some other methods prefer to non-linear mapping functions (e.g., \cite{Socher2013zero} uses a two-layer neural network).

It is worth noting that some methods such as \cite{Shigeto2015hubness} and \cite{zhang2017learning} learn an inverse mapping function $m^{-1}$ from the class semantic vector to the input.
The class of $x'$ is then searched in the input space, i.e., calculating $\arg \min_{y \in \mathcal{Y}_u} d(x', m^{-1}(h(y)))$.
This is believed to be able to release the hubness problem encountered in nearest neighbor search.
Besides, there are some methods that project the class semantic vector and the input to an intermediate space \cite{YangH2014multi,lei2015predicting}. 

\subsubsection{Generative Model Based}

These methods leverage a generative model (e.g., Generative Adversarial Network (GAN) \cite{goodfellow2014generative}) to synthesize training data for unseen classes conditioned on their semantic vectors.
Considering those methods using GAN, given the training data $\mathcal{D}_{tr}$, they learn a conditional generator $G:(h(y),v) \rightarrow \hat{x}$ which takes random Gaussian noise $v$ and the class semantic vector $h(y)$ as its input and outputs the synthetic sample $\hat{x}$ of class $y$, 
and at the same time, a discriminator $D: (x, \hat{x}) \rightarrow [0,1]$ is trained to distinguish the generated sample from the real sample.
Once the generator $G$ is trained to be able to synthesize plausible samples for seen classes (i.e., the synthesized and real samples cannot be distinguished by $D$), 
it is used to generate samples $\hat{x}_u$ for each unseen class $y_u$ in $\mathcal{Y}_u$ via its semantic vector $h(y_u)$.
This transforms the (generalized) ZSL problem into a normal supervised learning problem.

To improve the quality of generated samples, some works encourage the generator to generate samples that statistically match the distribution of real samples;
for example, \cite{qin2020generative} proposes a pivot regularization to enforce the mean of features of the generated samples to be that of the real samples.
Some other works encourage the generated samples to preserve the inter-class relationship indicated by the class semantic vectors;
for example, \cite{felix2018multi} develops a cycle consistency loss to enforce the generated samples to reconstruct their classes' semantic vectors.
%
Besides, more GAN variants such as 
StackGAN \cite{pandey2020stacked}, other generative models such as Variational Autoencoder (VAE) \cite{kingma2013auto}, and the combination of different generative models such as VAE and GAN \cite{xian2019f} have been explored.

\subsubsection{Graph Neural Network Based}\label{sec:gnnb}
These methods are mainly developed for graph-structured external knowledge where each class is aligned with one graph node and each inter-class relationship is represented by one graph edge.
Given the external knowledge graph, its nodes' states (semantic vectors) are initialized by e.g., word embeddings and multi-hot encodings, 
and a Graph Neural Network (GNN) is then applied to learn the nodes' semantic vectors with their relationships to the neighbours encoded.
Briefly, in every layer of the GNN, a propagation function is learned to update the state of each node by aggregating the features from its adjacent nodes in the graph, denoted as $p: h_u^l \rightarrow h_u^{l+1}$, where $h_u^l$ denotes the hidden state of node $u$ at the $l^{\text{th}}$ layer.
In the training, the classifiers of seen classes learned from $\mathcal{D}_{tr}$ are used to train the GNN, while in prediction, the GNN is used to calculate the classifiers of unseen classes.

Some methods implement the information propagation via convolutional operations \cite{kipf2016gcn}; for example, \cite{wang2018zero} uses a Graph Convolutional Neural Network for the external knowledge from WordNet \cite{miller1995wordnet}, while \cite{kampffmeyer2019rethinking} uses fewer convolutional layers but one additional dense connection layer to propagate features towards distant nodes for the same graph.
To enhance the information propagation, \cite{geng2020explainable} propose weighted aggregation to emphasize those more important adjacent nodes;
\cite{lee2018multi} designs 
multiple relation-specific functions
for the Graph Gated Neural Network to 
learn the propagation weights which control the information exchange between nodes.

In the above methods, GNNs are directly used to predict the classifiers of unseen classes.
Actually GNNs can also be used to embed the graph semantics (inter-class relationships) and calculate the semantic vectors $h(y)$ \cite{geng2021ontozsl,roy2020improving}.
With these class semantic vectors, different ZSL methods including the generative model based and the mapping function based can be further applied.
It is worth noting that for each class, multiple semantic vectors which may be calculated from different external knowledge resources, can be considered by simple concatenation or weighted combination \cite{roy2020improving}.


\section{External Knowledge}\label{sec:ek}
Different kinds of external knowledge have been explored to build the relationship between the seen and unseen classes. 
In this survey we divide them into four categories: text, attribute, KG, rule \& ontology, according to their characteristics, expressivity and the semantic encoding approaches.
See Table \ref{table:cek} for a brief summary.
In the remainder of this section, we will review the ZSL work for each category.
Note some work use more than one kind external knowledge.
This is as expected because different external knowledge often have different semantics and thus are complementary to each other.

\begin{table*}[t]
\footnotesize{
\centering
\renewcommand{\arraystretch}{1.4}
\begin{tabular}[t]{m{1.2cm}<{\centering}|m{6.3cm}<{\centering}|m{2.3cm}<{\centering}<{\centering}|m{1.3cm}<{\centering}|m{4.5cm}<{\centering}}\hline
 \textbf{Category} &\textbf{Description} &\textbf{Embedding} & \textbf{Semantic Richness}  & \textbf{Summary} \\ \hline
Text & Unstructured text that describe the classes, such as class names, phrases, sentences and documents & Word embedding, text feature learning  & Weak & Very easy to access; words are often ambiguous; long text is usually noisy  \\ \hline
Attribute & Semi-structured class properties with categorical, boolean or real values, such as  annotations that describe object visual characteristics & Vectors with binary or numeric values  & Medium  & Attributes by manual annotation are accurate but very costly  \\ \hline
KG & Multi-relation graph composed of entities aligned with the classes, other entities and their relationships such as the subsumption and the relational facts & Decomposition-based, translation-based, GNNs  & High  & KGs can also encompass the text and attribute external knowledge; some open KGs can be used \\ \hline
Ontology \& Rule & Logical relationships between the classes (and other concepts), such as the subsumption, the quantification constraints and the composition & Ontology embedding, materialization embedding 
&Very high  & Ontologies include KGs (as the fact parts) and can encompass the text and attributes; construction of logics relies on domain knowledge \\ \hline
\end{tabular}
\caption{\footnotesize A Brief Summary of The External Knowledge
}\label{table:cek}
}
\end{table*}

\subsection{Text}
Text external knowledge refer to unstructured textual information of the classes, such as their names, definitions and descriptions. 
They vary from words and phrases to long text such as sentences and documents.
Here are some typical examples in different ZSL tasks.
In image classification, \cite{norouzi2014zero},  \cite{Socher2013zero} and \cite{frome2013devise} utilize the class names and their word embeddings to address the unseen classes; \cite{elhoseiny2013write} and \cite{qiao2016less} prefer to existing class sentence descriptions from encyclopedia entries (articles); \cite{reed2016learning} collects more fine-grained and compact visual description sentences via crowdsourcing by the Amazon Mechanical Turk (AMT) platform.
In KG link prediction, \cite{qin2020generative} utilizes relation sentence descriptions from the KG itself for addressing  the unseen relations.
In entity extraction (a.k.a entity linking), \cite{logeswaran2019zero} deals with the unseen entities in a new specialized domain by using the entities' encyclopedia documents.

To encode the semantics of a class name, one approach is directly using its words' vectors by a language model or word embedding model that has been trained by a general purpose or domain specific corpus.
However, this makes the two tasks -- class semantic encoding and prediction model training detached with no interaction between them.
A coupled approach is jointly learning the prediction model and the class semantic encoding.
Note both can be pre-trained independently for high efficiency.
One representative method is DeViSE where a skip-gram word embedding model and an image classifier are fine-tuned jointly \cite{frome2013devise}.

Different from class names, textual class descriptions such as sentences and documents contain more yet noisy information. 
To suppress the noise, some additional feature learning and selection over the text (or text embedding) has been considered.
Among the aforementioned works, \cite{elhoseiny2013write} and \cite{qin2020generative} extract features from the text by the TF-IDF algorithm through which the vectors of some critical words get more weights; 
\cite{qiao2016less} initially encode the class descriptions into simple bag-of-words vectors, and then jointly learns text features and the image classifier; 
\cite{reed2016learning} also jointly learns text features and the image classifier, but considers both word-level and character-level text features.

In summary, the text external knowledge are easy to access for common ZSL tasks.
They can be extracted from not only the data of the ZSL tasks themselves but also encyclopedias, Web pages and other online resources. 
However, they are often noisy with irrelevant words and the words are always ambiguous.
They have limited expressivity on the semantics and cannot accurately express those fine-grained, logical or quantified inter-class relationships.

\subsection{Attribute}
Attribute external knowledge refer to those class properties with categorical, boolean or real values, which are often organized as key-value pairs.
They were originally explored in zero-shot object recognition, where the attributes are used to annotate visual characteristics such as object colours and shapes.
The simplest attributes are those binary annotations; for example, ``furry'' and ``striped'' indicate whether an animal looks furry and striped, respectively, in animal recognition \cite{lampert2009learning,lampert2013attribute}; while the annotation ``has wheel'' is used in recognizing vehicles \cite{farhadi2009describing}.
Relative attributes which enable comparing the degree of each attribute between classes (e.g., ``bears are furrier than giraffes'') \cite{parikh2011relative} are more expressive.
Another kinds of more expressive visual attributes are those associated with real values for quantifying the degree.
One typical example is the animal image classification benchmark named Animals with Attributes (AWA) \cite{lampert2013attribute,xian2018zero}.
Besides the visual tasks, the attribute external knowledge can also be applied in zero-shot graph link prediction.
\cite{hao2020inductive} utilizes the node attributes with categorical values to address the prediction involving unseen nodes which have no connection to the graph.

All the binary attributes of each class can be encoded into a multi-hot vector by associating one slot to one attribute and setting it to $1$ if the attribute is positive and to $0$ otherwise.
This can be extended to attributes with numeric values by filling their slots with their values.
For attributes with categorical values, each categorical value can be transformed into an integer.
In utilizing the attributes in image classification, the attribute vector can be directly used as the class semantic vector (i.e., $h(y)$), and then the mapping function based or the generative model based methods can be applied.
This kind of methods are also known as \textit{indirect attribute prediction} according to \cite{lampert2013attribute}.
In contrast, the other kind of methods are called \textit{direct attribute prediction} where the attributes of each testing sample are directly predicted.
The predicted attributes of a testing sample are  then used to determine the sample's class as the candidate that has the most similar attributes (e.g., \cite{parikh2011relative}).

In comparison with the text, the attribute external knowledge have higher expressivity with less noise and ambiguation.
They can even indirectly represent some quantified relationships between classes.
However, high quality attributes such as image annotations are often not available for a new task, while human annotations, which are often voted by multiple volunteers or even exerts, are very costly.
Therefore, in ZSL tasks beyond CV, they have been rarely explored.
In KG link prediction, entities and relations do not always have attributes while the existing attributes are often sparse.

\subsection{Knowledge Graph}
Another form of external knowledge for ZSL tasks is graph   knowledge represented as facts in RDF triple\footnote{\url{https://www.w3.org/TR/rdf11-concepts/}}, where seen and unseen classes are usually represented by  KG entities.
Considering animal image classification, a KG can express kinds of semantics for the inter-animal relationship, such as the animal taxonomy (e.g., \textit{cheetah} and \textit{jaguar} are species of \textit{big cats}) and the animal habitats (e.g., \textit{jaguar}  lives in \textit{forest} or \textit{swamp}; \textit{cheetah} lives in \textit{open area}).  
The exploration of KGs now mostly lies in ZSL tasks in CV, and most studies prefer to existing KGs that are open online. 
 \cite{wang2018zero}, \cite{lee2018multi} and \cite{kampffmeyer2019rethinking} adopt WordNet which includes semantic relationships such as synonyms and hyponyms.
\cite{wang2018zero} also studies NELL (a general KG extracted from the Web \cite{carlson2010toward}) for ZSL image classification.
\cite{roy2020improving} adopts the common sense KG named ConceptNet \cite{speer2017conceptnet}. 
\cite{geng2020explainable} constructs task-specific KGs using knowledge from both WordNet and DBpedia (a KG extracted from Wikipedia \cite{auer2007dbpedia})
while \cite{gao2019know} further considers ConceptNet besides WordNet and DBpedia.

As some   KGs, such as WordNet and ConceptNet, are often very large with much irrelevant knowledge, a task-specific KG is often constructed by knowledge extraction and integration.
To extract relevant knowledge, the task-specific data such as the class names are matched with KG entities either using some existing associations (e.g., ImageNet classes are matched with WordNet entities \cite{deng2009imagenet}) or by some mapping methods such as string matching.
With the constructed KG, the class semantic vectors  can then be learned by a KG embedding method which can be either GNN variants such as GCN by \cite{wang2018zero}, Relational GCN by \cite{roy2020improving} and Attentive GCN by \cite{geng2020explainable}, or translation based or factorization based KG embedding models such as TransE and DistMult. See \cite{wang2017knowledge} for a survey on KG embedding.

KG is more expressive than both text and attribute.
Besides the relational graph, a KG can represent and incorporate the text and attribute external knowledge as well. 
Text and attributes with real values can be represented by literals with data properties (such as \textit{obo:abstract}, \textit{rdfs:label} and \textit{dbo:populationMetro} in DBpedia), while attributes with binary or categorical values can be either represented as literals with (new) data properties or transformed into relational facts by creating new relations and new entities for the values \cite{geng2020explainable}.
A KG and its literals can be jointly embedded by some literal-aware KG embedding methods such as DKRL which supports entity descriptions \cite{xie2016representation}.
See \cite{gesese2020survey} for a survey.

On the other hand, KGs with suitable knowledge are not always available for a new real word ZSL task.
Although many public KGs such as ConceptNet and WordNet are very large, their contents are still incomplete and biased with a limited coverage for some specific domains.
Thus extracting, matching and curating knowledge from external resources and the task itself becomes a key challenge for KG-based ZSL.

\subsection{Rule \& Ontology}
Rule and ontology can express additional logical relationships between seen and unseen classes.
\cite{li2020logic} addresses the zero-shot relation extraction problem by using Horn rules to describe an unseen relation with some seen relations. 
For example, the unseen relation \textit{nominated\_for} is defined as the composition of two seen relations,
namely, $nominated\_for(x,z) \Leftarrow award\_received(x,y) \land winner(y,z)$.
Briefly this work matches all the candidate relations with Wikidata relations, construct a KG from Wikidata, learns the KG embeddings, and finally computes each unseen relation's semantic vector with the KG embeddings of its own and its compositional relations.
\cite{rocktaschel2015injecting} injects background knowledge on the relations in form of first-order formulae (e.g., $\forall x,y:daughter\_of(x,y) \Rightarrow has\_parents(x,y)$)
into relation embeddings to augment relation extraction and address unseen relations.
Three methods are considered to inject the rules: symbolic reasoning for ground atoms before or after learning the embeddings, jointly learning the embeddings with a loss term for the formulae.

An ontology is to represent and exchange knowledge such as names and definitions of the concepts and relations, their relationships, annotations, properties, etc.
The most common inter-concept or inter-relation relationship is the subsumption through which an ontology can represent the taxonomy. 
Ontologies can also define constraints and meta data (schema) of the concepts and relations such as the concept existential quantifier and the relation domain and range.
Besides, ontologies by Web Ontology Language (OWL)\footnote{\url{https://www.w3.org/TR/owl-features/}} are able to represent quite a few logical relationships. 
Considering the above rule about \textit{nominated\_for}, an OWL ontology can represent it by defining a complex relation with atom relations, i.e., $nominated\_for \equiv award\_received \circ winner$ (a.k.a. \textit{role composition}).
\cite{geng2021ontozsl} studies zero-shot animal image classification and KG link prediction for unseen relations with two ontologies of RDF Schema (RDFS), respectively.
The former ontology encompasses an animal taxonomy, annotations and text descriptions; while the later ontology encompasses the relations, concepts, their hierarchies (e.g., $radiostation\_in\_city \sqsubseteq 
office\_in\_city$), the relation domain and range (e.g., $radiostation\_in\_city$ should be followed by instances that belong to \textit{city}).
The semantic vectors of the animal classes (which are aligned to ontology concepts) and the relations are then leaned by literal aware KG embedding (RDFS ontologies can be directly transformed into RDF KGs), and a generative model based ZSL method is eventually applied.
\cite{chen2020ontology} explores OWL ontology for augmenting ZSL with animal image classification.
The ontology includes an animal taxonomy and complex concepts defined by the composition of atom concepts and existential quantifiers (e.g., $Killer\_Whale \equiv Toothed\_Whale \sqcap \exists hasTexture.Patches \sqcap \cdots$), following the EL fragment of OWL 2 which can be embedded by geometric methods such as \cite{kulmanov2019embeddings} or word embedding based methods such as OWL2Vec* \cite{chen2020owl2vec}.


\section{Application and Benchmark}\label{sec:ab}

\begin{table*}[t]
\footnotesize{
\centering
\renewcommand{\arraystretch}{1.4}
\begin{tabular}[t]{m{1.6cm}<{\centering}|m{1.4cm}<{\centering}|m{6.2cm}<{\centering}<{\centering}|m{2.9cm}<{\centering}|m{3.3cm}<{\centering}}\hline
 \textbf{Tasks} &\textbf{Names} &\textbf{Information (Splitting)} & \textbf{External Knowledge}  & \textbf{Sources}   \\ \hline
\multirow{4}{*}{\tabincell{\\  \\Image\\Classification}} & \shortstack{aPY\\ /CUB\\ /SUN\\ /AWA1\\ /AWA2}  & 5 benchmarks with 20/150/645/40/40 seen classes, 12/50/72/10/10 unseen classes and 15329/11788/14340/37475/37322 images; note classes of CUB and SUN are fine-grained & 64/312/102/85/85 visual annotations (attributes) and class names & The latest (cleaned) version can be found in \cite{xian2018zero} \\ \cline{2-5}
 & ImageNet & 1K classes of ImageNet 2012 as the seen; classes that are 2/3-hops away, or all the other ImageNet classes (21K) as the unseen & The WordNet KG & Splits proposed by \cite{frome2013devise} and \cite{wang2018zero}  \\ \cline{2-5}
 & ImNet-A & 28 seen classes (37,800 images) and 52 unseen classes (39,523 images) & \multirow{2}{*}{\shortstack{Ontologies with \\ animal taxonomies, \\ names, attributes and \\ textual descriptions}} & \multirow{2}{*}{\shortstack{Images from ImageNet;\\ splits and external \\  knowledge by \\ \cite{geng2021ontozsl}}}  \\ \cline{2-3}
 & ImNet-O & 10 seen classes (13,407 images) and 25 unseen classes (25,954 images)  &  &   \\ \hline
Visual Question Answering & VQA dataset & 2951/811 seen/unseen objects; 
a testing question has at least one unseen object; 
20,472 images, 614,164 questions, 10 answers per question;
&Object description from Wikipedia and books on the Web & The version with splits and object description by \cite{ramakrishnan2017empirical} \\ \hline
\multirow{4}{*}{\tabincell{\\Link\\Prediction}} & NELL-ZS  & KG from NELL with 65,567 entities and 188,392 triples; 149/32 seen/unseen relations & \multirow{2}{*}{\shortstack{Textual description, \\names and ontologies \\ (e.g., relation hierarchy \\ relation domain/range)}} & \multirow{2}{*}{\shortstack{Original version with text \\ by \cite{qin2020generative}; new \\ version with ontologies \\by  \cite{geng2021ontozsl}}} \\ \cline{2-3}
 & Wikidata-ZS & KG from Wikidata with 605,812 entities and 724,967 triples; 489/48 seen/unseen relations & & \\ \cline{2-5}
 & FB20k & KG from Freebase with 1,341 relations and 14,904/5,019 seen/unseen entities & Textual descriptions & \cite{xie2016representation}\\ \cline{2-5}
 & DBpedia50k & KG from DBpedia with 654 relations and 46,264/3,636 seen/unseen entities & Textual descriptions & \cite{shi2018open}  \\ \hline
 %
\multirow{2}{*}{\tabincell{Text\\Classification}}  & DBpedia-Wikipedia  &14 non-overlapping DBpedia classes with 45,000 docs (samples) from Wikipedia; 11/3 (or 7/7) seen/unseen classes & \multirow{2}{*}{\shortstack{Class names, \\ textual descriptions,\\ class hierarchy, \\ the ConceptNet KG}} &\multirow{2}{*}{\shortstack{The version with \\ external knowledge \\ and splits proposed \\ by \cite{zhang2019integrating}}} \\ \cline{2-3}
 &20news-group & 20 topics each of which has around 1,000 docs; 15/5 (or 10/10) seen/unseen classes & & \\ \hline
\multirow{2}{*}{\tabincell{\\Relation\\Extraction}} & Reading Comprehension & 840,000 samples with $10$ folds of train/dev/test; 84/12 train/dev relations as the seen; 12 test relations as the unseen & Relation names, question templates (textual explanations) & \cite{levy2017zero} \\ \cline{2-5}
 & Expanded SWDE & 21 English websites; each website has 400 to 2,000 pages; 18/14/13 relations for the Movie/NBA/University vertical; two verticals the seen, one vertical as the unseen  & Relation names, web page text fields & \cite{lockard2020zeroshotceres} \\ \hline
\end{tabular}
\caption{\footnotesize A Summary of Some Open ZSL Benchmarks with Novel External Knowledge
}\label{table:benchmark}
}
\end{table*}

This section introduces popular ZSL applications.
Table \ref{table:benchmark} briefly summarizes some open benchmarks with novel external knowledge. 
Those normal supervised learning benchmarks that are used to evaluate ZSL by new splits are ignored.

ZSL has been widely investigated for image classification for both general purpose tasks such as those using ImageNet and domain-specific tasks such as mammal classification (cf. the AWA1 and AWA2 benchmarks \cite{lampert2013attribute,xian2018zero}) and fine-grained bird classification (cf. the CUB benchmark \cite{welinder2010caltech}).
Almost all these tasks allow to use class names while more textual information sometimes is accessed online.
Manual visual annotations (attributes) are widely studied, but are mostly limited to domain-specific tasks with a small to medium number of classes. 
KGs and ontologies are often studied for tasks whose classes have been aligned to entities of some KG; for example, ImageNet classes are aligned to WordNet entities.
Another CV task is visual question answering (VQA) which aims to predict the answer of a natural language question according to the content of an image.
ZSL in VQA has several different definitions. 
\cite{chen2020ontology} defines ZSL as a setting where some testing answers have never appeared in training, while \cite{ramakrishnan2017empirical} defines ZSL as a setting where the image or the question contains some novel object (e.g., ‘Is the dog black?’ where ‘dog’ has never appeared in training).
\cite{teney2016zero} has a more tolerant definition which regards a question as unseen if there is at least one novel word in this question or in its answer.

ZSL has also been studied in NLP for e.g., text classification and open information extraction.
The most common external knowledge for text classification are class names (with word embeddings pre-trained by an external corpus) and text descriptions \cite{srivastava2018zero}. 
The class hierarchy and KG are found to be investigated by one study, i.e., \cite{zhang2019integrating}.
Most current evaluation re-uses the existing benchmarks for e.g., topic categorization and emotion detection with new seen and unseen splits \cite{yin2019benchmarking}.
In information extraction, ZSL aims at addressing unseen relations and entities. 
One popular direction is exploring kinds of textual information such as entity descriptions \cite{logeswaran2019zero}, relation web pages \cite{lockard2020zeroshotceres} and question template-based relation explanations \cite{levy2017zero}. 
KGs and rules have been explored (e.g., \cite{li2020logic}) but not widely, with no open benchmarks.
Different from CV tasks, ZSL text classification and open information extraction have rarely considered attributes.

Recently ZSL has been applied to address emerging entities and relations in KG embedding and link prediction.
Note we exclude those new entities or relations that have a few links to the existing KG as they lead to some training samples.
There are only a small number of relevant papers, but the external knowledge of text descriptions and ontological schemas have been explored (cf.  \cite{qin2020generative}, \cite{xie2016representation} and \cite{geng2021ontozsl}, respectively).
Besides, ZSL has also been applied in other prediction tasks involving KGs such as question answering \cite{banerjee2020self}.

\section{Conclusion and Discussion}\label{sec:con}
This paper presents a literature review for ZSL mainly from the perspective of external knowledge. We   briefly reviewed the ZSL definitions, technical solutions, applications and open benchmarks.
We divided the external knowledge into text, attribute, KG, and ontology  \& rule.
For each category, we introduced the characteristics, presented how they are utilized and summarized the advantages and disadvantages.
We analyzed those typical papers for the text and attribute external knowledge due to the space limitation, 
while giving a rather complete review to those ZSL works using KGs, rules and ontologies as  external knowledge.

\vspace{0.05cm}
\noindent\textbf{Trend Analysis.}
After ZSL was proposed in around 2009, it was widely investigated for image classification at the beginning with attributes (e.g., \cite{lampert2009learning,farhadi2009describing,parikh2011relative}) and then with kinds of text and text-based latent embeddings (e.g., DeViSE \cite{frome2013devise} and SAE \cite{kodirov2017semantic}).
KG for ZSL can be traced back to \cite{palatucci2009zero} which uses two small KGs containing word attributes and word co-occurrence relationships respectively.  
However, KG and ontology for ZSL were then not studied until in recent four years when general purpose KGs, such as NELL and Wikidata,   became popular (e.g., \cite{lee2018multi,wang2018zero,geng2020explainable}). 
Due to higher expressivity, compatibility to text and attribute, more and more resources and embedding techniques, we believe KGs and ontologies will continue to play an increasingly important role in ZSL.
Meanwhile, ZSL has been extended from CV to other domains such as NLP, KG construction and completion in recent five years.
This trend will continue as the real world contexts often violate the normal supervised learning setting with emerging classes.

\vspace{0.05cm}
\noindent\textbf{Few-shot \& Transfer Learning.}
ZSL is highly relevant to another two widely investigated directions: \textit{few-shot learning} (FSL) where one or only a few labeled samples are available for some classes, 
and \textit{transfer learning} (TL) which transfers data or model from one domain to another different yet related domain.
Since some labeled samples are given, methods for FSL focus more on generalizing from these samples by e.g., meta learning and TL algorithms, and fewer FSL works study the usage of external knowledge
\cite{wang2020generalizing}.
However, we argue that combining the available samples and the external knowledge would be a promising direction in many tasks (e.g., \cite{rios2018few} utilizes the class label structure for few-shot text classification
).
On the other hand, TL studies have led to   a few effective algorithms for sharing samples and models, most of which can be applied in FSL, ZSL and other sample shortage settings.
External knowledge such as those on the domains and features can be used to select transferable models and samples, avoid negative transfer and explain the transfer (e.g., \cite{chen2018knowledge} uses the ontology knowledge on airlines, airports and so on to explain the transferability of neural network features for flight delay forecasting). 
These ZSL, FSL and TL studies which explore symbolic knowledge represented by KG, ontology and logical rule form a new and interesting neural-symbolic paradigm which injects background and human knowledge in learning for addressing ML sample shortage challenges.


\bibliographystyle{named}
\bibliography{zsl}

\balance

\end{document}